\documentclass{article} 
\usepackage{arxiv,times}

\usepackage{microtype}
\usepackage{graphicx}
\usepackage{subfigure}
\usepackage{booktabs} 
\usepackage{amsmath,amsthm,amsfonts,thmtools}
\usepackage{amssymb}
\usepackage{multirow}
\usepackage{xcolor}
\usepackage{natbib}

\usepackage{mdframed}

\usepackage{euscript}

\usepackage{hyperref}

\newcommand{\bX}{{\bf X}}

\newcommand{\bW}{{\bf W}}

\newcommand{\bb}{{\bf b}}

\newcommand{\bA}{{\bf A}}

\newcommand{\bY}{{\bf Y}}
\newcommand{\bF}{{\bf F}}

\newcommand{\bc}{{\bf c}}

\newcommand{\cG}{\mathcal{G}}
\newcommand{\cN}{\mathcal{N}}
\newcommand{\cV}{\mathcal{V}}
\newcommand{\cE}{\EuScript{E}}

\newcommand{\Dt}{{\Delta t}}

\newcommand{\fref}[1] {Fig.~\ref{#1}}

\newcommand{\ident}{\mathrm{I}}

\newcommand{\bD}{{\bf D}}

\newmdtheoremenv{theo}{Theorem}
\newmdtheoremenv{defi}[theo]{Definition}

\newtheorem{theorem}{Theorem}[section]

\newtheorem{remark}[theorem]{Remark}

\title{\resizebox{\textwidth}{!}{A Survey on Oversmoothing in Graph Neural Networks}}

\author{T. Konstantin Rusch\\
ETH Zurich\\
\texttt{trusch@ethz.ch}
\\
\And Michael M. Bronstein \\
University of Oxford \\
\And
Siddhartha Mishra \\
ETH Zurich\\
}

\date{}
\begin{document}

\maketitle

\begin{abstract}
Node features of graph neural networks (GNNs) tend to become more similar with the increase of the network depth. This effect is known as {\em over-smoothing}, which we axiomatically define as the exponential convergence of suitable similarity measures on the node features. Our definition unifies previous approaches and gives rise to new quantitative measures of over-smoothing. Moreover, we empirically demonstrate this behavior for several over-smoothing measures on different graphs (small-, medium-, and large-scale). We also review several approaches for mitigating over-smoothing and empirically test their effectiveness on real-world graph datasets. 
Through illustrative examples, we demonstrate that mitigating over-smoothing is a necessary but not sufficient condition for building deep GNNs that are expressive on a wide range of graph learning tasks. Finally, we extend our definition of over-smoothing to the rapidly emerging field of continuous-time GNNs.
\end{abstract}

\section{Introduction}
Graph Neural Networks (GNNs) \citep{sperduti1994encoding,goller1996learning,sperduti1997supervised,frasconi1998general,gori2005new,scarselli2008graph,bruna2013spectral,chebnet,gcn,MoNet,mpnn} have emerged as a powerful tool for learning on relational and interaction data. These models have been successfully applied on a variety of different tasks, e.g. in computer vision and graphics \cite{MoNet}, recommender systems \cite{ying2018graph}, transportation \cite{derrowpinion2021traffic}, computational chemistry \citep{mpnn},  drug discovery \cite{gaudelet2021utilizing}, particle physics \citep{shlomi2020graph}, and analysis of social networks (see \citet{zhou,gdlbook} for additional applications).

The number of layers in a neural network (referred to as ``depth'' and giving the name to the entire field of ``deep learning'') is often considered to be crucial for its performance on real-world tasks. For example, convolutional neural networks (CNNs) used in computer vision, often use tens or even hundreds of layers. 
In contrast, most GNNs encountered in applications are relatively shallow and often have just few layers. 
This is related to several issues impairing the performance of deep GNNs in realistic graph-learning settings: graph bottlenecks  \citep{alon2020bottleneck}, over-squashing \citep{topping2021understanding,egp}, and over-smoothing \citep{li2018deeper,os1,os2}. In this article we focus on the over-smoothing phenomenon, which \emph{loosely} refers to the exponential convergence of all node features towards the same constant value as the number of layers in the GNN increases. While it has been shown that small amounts of smoothing are desirable for regression and classification tasks \citep{kerivennot}, excessive smoothing (or `over-smoothing') results in convergence to a non-informative limit. Besides being a key limitation in the development of deep multi-layer GNNs, over-smoothing can also severely impact the ability of GNNs to handle {\em heterophilic} graphs \citep{syn_cora}, in which node labels tend to differ from the labels of the neighbors and thus long-term interactions have to be learned.

Recent literature has focused on precisely defining over-smoothing through measures of node feature similarities such as the the graph Dirichlet  \citep{graphcon,dirich_first,pairnorm,energetic_gnn}, cosine similarity \citep{mad}, and other related similarity scores \citep{group_ratio}. With the abundance of such measures, however, there is currently still a conceptual gap in a  general definition of over-smoothing that would provide a unification of existing approaches. Moreover, previous work mostly measures the similarity of node features and does not explicitly consider the rate of convergence of over-smoothing measures with respect to an increasing number of GNN layers.  

In this article, we aim to unify several recent approaches and define over-smoothing in a formal and tractable manner through an axiomatic construction.  
Through our definition, we rule out problematic measures such as the Mean Average Distance that does not provide a sufficient condition for over-smoothing. 
We then review several approaches to mitigate over-smoothing and provide their extensive empirical evaluation. These empirical studies lead to the insight that meaningfully solving over-smoothing in deep GNNs is more elaborate than simply forcing the node features not to  converge towards the same node value when the number of layers is increased. Rather, there needs to be a subtle balance between the expressive power of the deep GNN and its ability to preserve the diversity of node features in the graph. Finally, we extend our definition to the rapidly emerging sub-field of continuous-time GNNs \citep{grand,blend,graphcon,g2,sheaf,grad_flow}. 

\section{Definition of over-smoothing}
Let $\mathcal{G}=(\mathcal{V},\mathcal{E}\subseteq \mathcal{V}\times \mathcal{V})$ be an undirected graph with $|\mathcal{V}|=v$ nodes and $|\mathcal{E}|=e$ edges (unordered pairs of nodes $\{ i,j \}$   denoted $i \sim j$). 
The \emph{$1$-neighborhood} of a node $i$ is denoted  
$
\cN_i =
\{j \in \cV : i\sim j \}
$.
Furthermore, each node $i$ is endowed with an $m$-dimensional feature vector $\mathbf{X}_i$; the node features are arranged into a $v\times m$ matrix 
$\bX = (\bX_{ik})$ with $i=1,\hdots, v$ and $k=1,\hdots, m$.

Message-Passing GNN (MPNN) updates the node features by performing several iterations of the form,
\begin{equation}
\label{eq:mp}
\bX^n = \sigma(\bF_{\theta^n}(\bX^{n-1},\cG)), \quad \forall n=1,\dots,N,
\end{equation}
where $\bF_{\theta^n}$ is a \emph{learnable} function  with parameters $\theta^n$, $\bX^n \in \mathbb{R}^{v \times m_n}$ are the $m$-dimensional hidden node features, and $\sigma$ is an element-wise non-linear activation function. Here, $n \geq 1$ denotes the $n$-th layer with  $n=0$ being the input layer and $N$ the total number of layers (depth). In particular, we consider local (1-neighborhood) coupling  of the form 
$(\mathbf{F}(\mathbf{X},\cG))_i = \mathbf{F}(\mathbf{X}_i,  \mathbf{X}_{j\in \cN_i})$ operating on the multiset of 1-neighbors of each node. Examples of such functions used in the graph machine learning literature \citep{gdlbook} include {\em graph convolutions} \citep{gcn} and {\em attentional message passing} \citep{gat}.

There exist a variety of different approaches to quantify over-smoothing in deep GNNs, e.g. measures based on the Dirichlet energy on graphs \citep{graphcon,dirich_first,pairnorm,energetic_gnn}, as well as measures based on the mean-average distance (MAD) \citep{mad,group_ratio}, and references therein. However, most previous approaches lack a formal definition of over-smoothing as well as provide approaches to measure over-smoothing which are not sufficient to quantify this issue. Thus, the aim of this survey is to establish a unified, rigorous, and tractable definition of over-smoothing, which we provide in the following.
\vspace{1em}
\begin{defi}[Over-smoothing]
\label{def:oversmoothing}
Let $\cG$ be an undirected, connected graph and $\bX^n \in \mathbb{R}^{v \times m}$ denote the $n$-th layer hidden features of an $N$-layer GNN defined on $\cG$. 
Moreover, we call $\mu:\mathbb{R}^{v \times m} \longrightarrow \mathbb{R}_{\geq 0}$ a \textbf{node-similarity measure} if it satisfies the following axioms:
\begin{enumerate}
    \item $\exists \bc \in \mathbb{R}^{m}$ with $\bX_i=\bc$ for all nodes $i \in \cV$ $\Leftrightarrow$ $\mu(\bX)=0$, ~for $\bX \in \mathbb{R}^{v \times m}$
    \item $\mu(\bX + \bY) \leq \mu(\bX) + \mu(\bY)$, ~for all $\bX,\bY \in \mathbb{R}^{v \times m}$
\end{enumerate}
We then define {\bf over-smoothing with respect to $\mu$} as the layer-wise exponential convergence of the node-similarity measure $\mu$ to zero, i.e.,
\begin{enumerate}
    \item[3.] $\mu(\bX^n) \leq C_1e^{-C_2n}$, for $n=0,\dots,N$ with some constants $C_1,C_2>0$.
\end{enumerate}
\end{defi}
Note that without loss of generality we assume that the node-similarity measure $\mu$ converges to zero (any node-similarity measure that converges towards a non-zero constant can easily be recast). Further remarks about Definition \ref{def:oversmoothing} are in order.

\begin{remark}
Condition 1 in Definition \ref{def:oversmoothing} simply formalizes the widely accepted notion that over-smoothing is caused by node features converging to a constant node vector whereas condition 3 provides a more stringent, quantitative measure of this convergence. Note that the triangle inequality or subadditivity (condition 2) rules out degenerate choices of similarity measures.
\end{remark}

\begin{remark}
Definition \ref{def:oversmoothing} only considers the case of connected graphs. However, this definition can be directly generalized to disconnected graphs, where we apply a node-similarity measure $\mu_S$ on every connected component $S \subseteq \cV$, and define the global similarity measure as the sum of the node-similarity measures on each connected component, i.e., $\mu = \sum_{S}\mu_S$. This way we ensure to cover the case of different connected components converging to different constant node values. 
\end{remark}

\section{Over-smoothing measures}
\label{sec:measures}
Existing approaches to measure over-smoothing in deep GNNs have mainly been based on concept of \emph{Dirichlet energy} on graphs,
\begin{equation}
\label{eq:dirichlet}
\cE(\bX^n) = \frac{1}{v}\sum_{i\in \cV} \sum_{j \in \cN_i} \|\bX^n_i - \bX^n_j \|^2_2,
\end{equation}
(note that instead of normalizing by $1/v$ we can equivalently normalize the terms inside the norm based on the node degrees $d_i$, i.e. $\|\frac{\bX^n_i}{\sqrt{1+d_i}} - \frac{\bX^n_j}{\sqrt{1+d_j}} \|^2_2$). It is straightforward to check that the measure, 
\begin{equation}
    \label{eq:dirsim}
    \mu(\bX^n) = \sqrt{\cE(\bX^n)},
\end{equation}
satisfies the conditions 1 and 2 in the definition \ref{def:oversmoothing} and thus, constitutes a bona fide node-similarity measure. Note that in the remainder of this article, we will refer to the square root of the Dirichlet energy simply as the Dirichlet energy.

In the literature, \emph{Mean Average Distance} (MAD),
\begin{equation}
\label{eq:mad}
\mu(\bX^n) = \frac{1}{v}\sum_{i\in \cV} \sum_{j \in \cN_i} 1 - \frac{{\bX^n_i}^\top\bX^n_j}{\|\bX^n_i\|\|\bX^n_j\|}.
\end{equation}
has often been suggested as a measure of over-smoothing. 
We see that in contrast to the Dirichlet energy, \emph{MAD is not a node-similarity measure}, as it does not fulfill condition 1 nor condition 2 of the over-smoothing definition \ref{def:oversmoothing}. In fact, MAD is always zero in the scalar case, where all node features share the same sign for each feature dimension. This makes MAD a very problematic measure for over-smoothing as $\mu(\bX)=0$ does not represent a sufficient condition for over-smoothing to happen. However, as we will see in the subsequent section, in the multi-dimensional case ($m>1$) MAD does converge exponentially to zero for increasing number of layers if the GNN over-smooths and thus fulfills condition 3 of the over-smoothing definition \ref{def:oversmoothing}. Therefore, we conclude that under careful considerations of the specific use-case, MAD may be used as a measure for over-smoothing. However, since the Dirichlet energy fulfills all three conditions of Definition \ref{def:oversmoothing} and is numerically more stable to compute, it should always be favored over MAD. 

It is natural to ask if there exist other measures that constitute a node-similarity measure as of Definition \ref{def:oversmoothing} and can thus be used to define over-smoothing. While the Dirichlet energy denotes a canonical choice in this context, there are other measures that can be used. For instance, instead of basing the Dirichlet energy in \eqref{eq:dirichlet} on the $L^2$ norm, any other $L^p$-norm ($p>1$) can be used.

\subsection{Empirical evaluation of different measures for over-smoothing}
\citet{graphcon} have empirically demonstrated the qualitative behavior described in Definition \ref{def:oversmoothing} on a $10\times10$ regular 2-dimensional grid with one-dimensional uniform random (hidden) node features. We extend this empirical study in two directions, first to higher dimensional node features and also to real-world graphs, namely Texas \citep{geom_gcn}, Cora \citep{cora}, and Cornell5 (Facebook 100 dataset).
Note that as mentioned above, the extension to higher dimensional node features is necessary in order to empirically evaluate MAD, as MAD is zero for any one-dimensional node features sharing the same sign.

\begin{figure}[ht!]
\begin{minipage}{1.\textwidth}
\includegraphics[width=.33\textwidth]{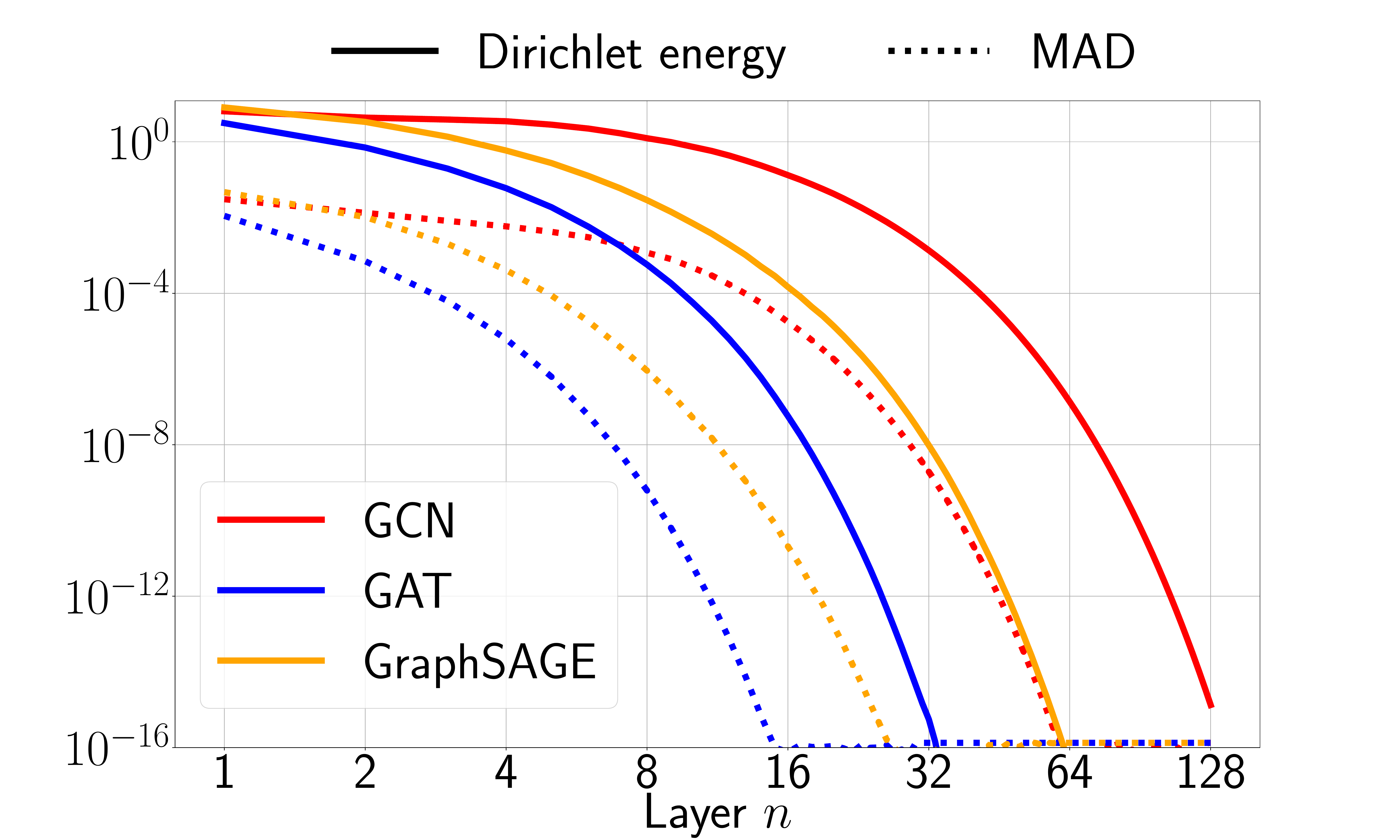}%
\includegraphics[width=.33\textwidth]{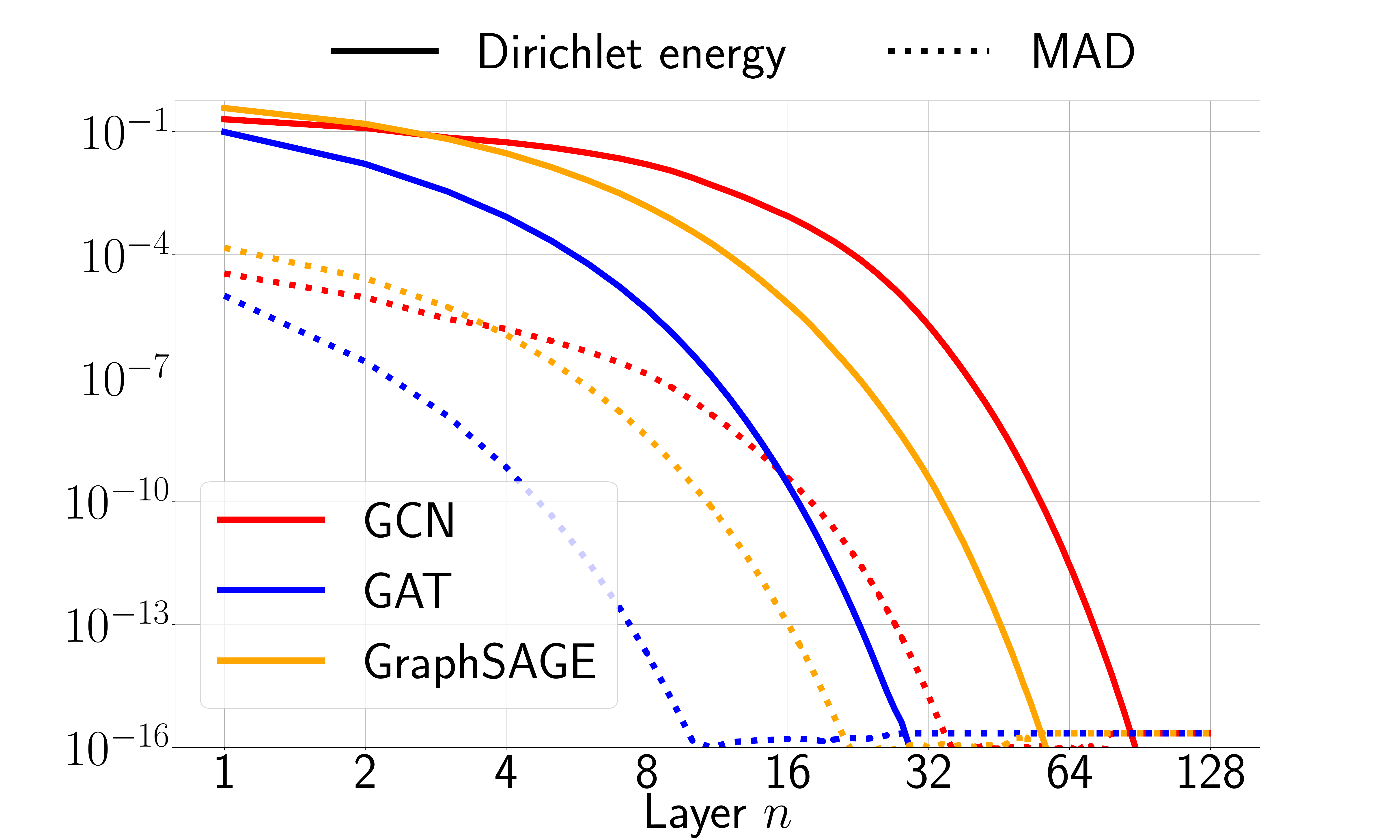}%
\includegraphics[width=.33\textwidth]{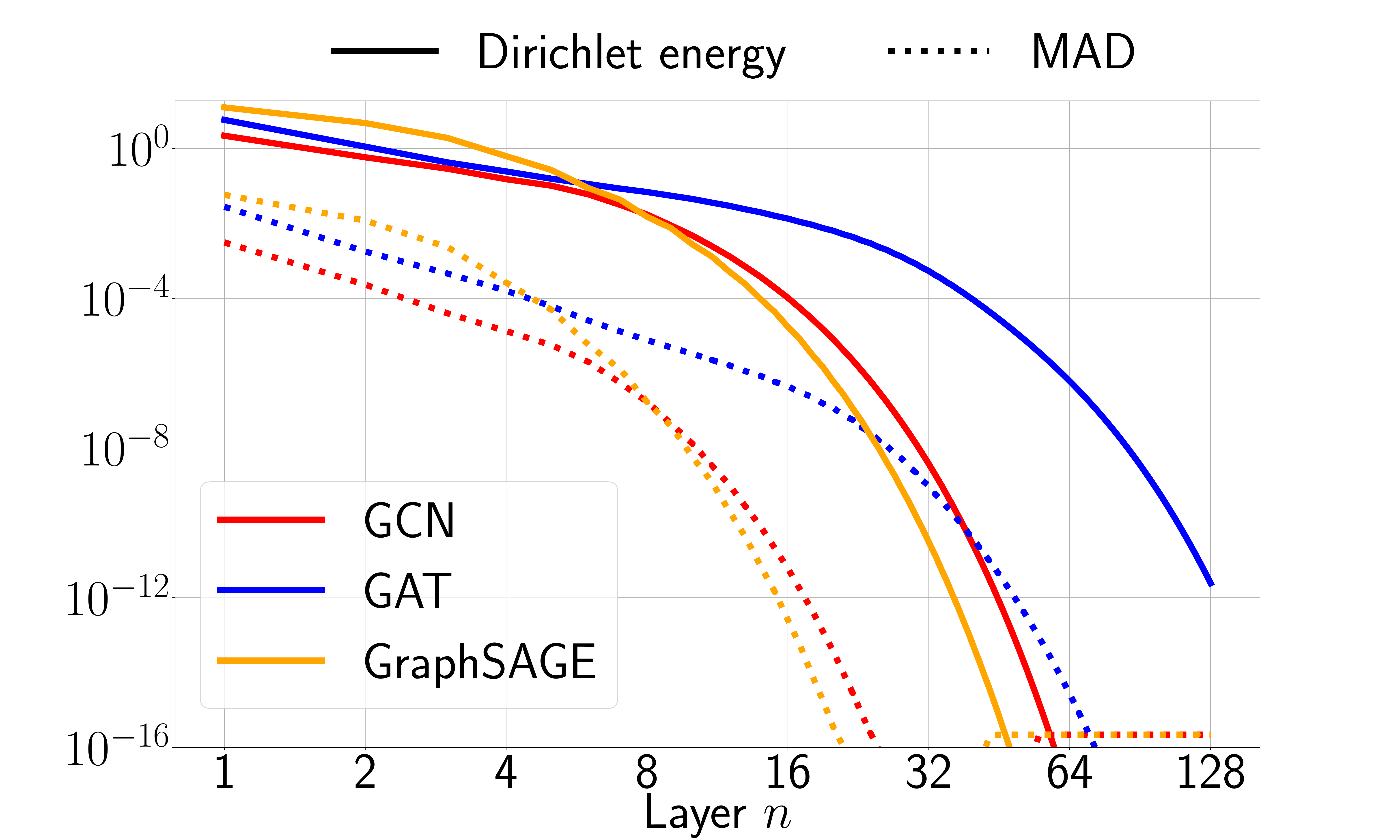}
\caption{Dirichlet energy and Mean Average Distance (MAD) of layer-wise node features $\bX^n$ propagated through a GAT, GCN and GraphSAGE for three different graph datasets, \textbf{(left)} small-scale Texas graph, \textbf{(middle)} medium-scale Cora citation network, (\textbf{right}) large-scale Facebook network (Cornell5).}
\label{fig:os_measures_plot}
\end{minipage}
\end{figure}

Since we are interested only in the dynamics of the Dirichlet energy and MAD associated with the propagation of node features through different GNN architectures, we omit the original input node features of the real-world graph dataset Cora and exchange them for standard normal random variables, i.e., $\bX_{jk} \sim \mathcal{N}(0,1)$ for all nodes $j$ and every feature $k$. In \fref{fig:os_measures_plot} we set the input and hidden dimension of the node features to $128$ and plot the (logarithm of) the Dirichlet energy \eqref{eq:dirichlet} and MAD \eqref{eq:mad}) of each layer's node features with respect to the (logarithm of) layer number for three popular GNN models, i.e.,  GCN, GAT and the GraphSAGE architecture of \cite{graphsage}. We can see that all three GNN architectures \emph{over-smooth}, with both layer-wise measures converging exponentially fast to zero for increasing number of layers. Moreover, we observe that this behavior is not just restricted to the structured and regular grid dataset of \cite{graphcon}, but the same behavior (i.e., exponential convergence of the measures with respect to increasing number of layers) can be seen on all the three real-world graph datasets considered here. 

It is important to emphasize the importance of the \textbf{exponential convergence} of the layer-wise over-smoothing measure $\mu$ to zero in Definition \ref{def:oversmoothing}. Algebraic convergence is not sufficient for the GNN to suffer from over-smoothing. This can be seen in \fref{fig:os_measures_plot}, where for instance the Dirichlet energy of GCN, GraphSAGE and GAT reach machine-precision zero after a maximum of $64$ layers, while for instance a linear convergence of the Dirichlet energy would still have a Dirichlet energy of around $1$ for an initial energy of around $100$, even after $128$ hidden layers.  

\section{Reducing over-smoothing}
\subsection{Methods}
Several methods to mitigate (or at least reduce) the effect of over-smoothing in deep GNNs have recently been proposed. While we do not discuss each individual approach, we highlight several recent methods in this context, all of which can be classified into one of the following classes.

\paragraph{Normalization and Regularization}
A proven way to reduce the over-smoothing effect in deep GNNs is to regularize the training procedure. This can be done either explicitly by penalizing deviations of over-smoothing measures during training or implicitly by normalizing the node feature embeddings and by adding noise to the optimization process.
An example of explicit regularization techniques can be found in Energetic Graph Neural Networks (EGNNs) \citep{energetic_gnn}, where the authors measure over-smoothing using the Dirichlet energy and propose to optimize a GNN within a constrained range of the underlying layer-wise Dirichlet energy.
DropEdge \citep{dropedge} on the other hand represents an example of implicit regularization by adding noise to the optimization process. This is done by randomly dropping edges of the underlying graph during training. Graph DropConnect (GDC) \citep{gdc} generalizes this approach by allowing the GNNs to draw different random masks for each channel and edge independently.
Another example of implicit regularization is PairNorm \citep{pairnorm}, where the pairwise distances are set to be constant throughout every layer in the deep GNN. This is obtained by performing the following normalization on the node features $\bX$ after each GNN layer,
\begin{equation}
    \begin{aligned}
        \hat{\bX}_i &= \bX_i - \frac{1}{v}\sum_{j=1}^v \bX_{j}, \\ \bX_i &= \frac{s\hat{\bX}_i}{\sqrt{\frac{1}{v}\sum_{j=1}^v \|\hat{\bX}_{j}\|_2^2}}, 
    \end{aligned}
\end{equation}
where $s>0$ is a hyperparameter.
Similarly, \citet{group_ratio} have suggested to normalize within groups of the same labeled nodes, leading to Differentiable Group Normalization (DGN). Moreover, \citet{nodenorm} have suggested to node-wise normalize each feature vector, yielding NodeNorm.

\paragraph{Change of GNN dynamics}
A rapidly emerging strategy to mitigate over-smoothing for deep GNNs is by qualitatively changing the (discrete or continuous) dynamics of the message-passing propagation. A recent example is the use of non-linear oscillators which are coupled through the graph structure yielding Graph-Coupled Oscillator Network (GraphCON) \citep{graphcon},
\begin{equation}
\begin{aligned}
\label{eq:graphCON}
    \bY^n &= \bY^{n-1} + \Dt [\sigma(\bF_{\theta^n}(\bX^{n-1},\cG)) - \gamma\bX^{n-1} - \alpha \bY^{n-1}], \\
    \bX^n &= \bX^{n-1} + \Dt\bY^n,
\end{aligned}
\end{equation}
where $\bY^n$ are auxiliary node features and $\Delta t>0$ denotes the time-step (usually set to $\Delta t =1$). The idea of this work is to exchange the diffusion-like dynamics of GCNs (and its variants) to that of non-linear oscillators, which can provably be guaranteed to have a Dirichlet energy that does not exponentially vanish (as of Definition \ref{def:oversmoothing}).
A similar approach has been taken in \cite{pde-gcn}, where the dynamics of a deep GCN is modelled as a wave-type partial differential equation (PDE) on graphs, yielding PDE-GCN. Another approach inspired by physical systems is Allen-Cahn Message Passing (ACMP) \citep{allen_cahn_gnn}, where the dynamics is constructed based on the Allen-Cahn equation modeling interacting particle system with attractive and repulsive forces. A related effort is the Gradient Flow Framework (GRAFF) \citep{grad_flow}, where the proposed GNN framework can be interpreted as attractive respectively repulsive forces between adjacent features.

A recent example in this direction, that is not directly inspired by physical systems, is that of Gradient Gating ({\bf G$^2$}) \citep{g2}, where a learnable node-wise early-stopping mechanism is realized through a gating function leveraging the graph-gradient,
\begin{equation}
\begin{aligned}
\label{eq:g2}
    \hat{\boldsymbol{\tau}}^n &= {\sigma}(\hat{\bF}_{\hat{\theta}^n}(\bX^{n-1},\cG)),\\
    \boldsymbol{\tau}^n_{ik} &=
    \tanh\left(\sum_{j\in\cN_i}|\hat{\boldsymbol{\tau}}^n_{jk} - \hat{\boldsymbol{\tau}}^n_{ik}|^p\right), \\
    \bX^n &=  (1 - \boldsymbol{\tau}^n)\odot \bX^{n-1} + \boldsymbol{\tau}^n \odot \sigma(\bF_{\theta^n}(\bX^{n-1},\cG)),
\end{aligned}
\end{equation}
with $p\geq0$. This mechanism slows down the message-passing propagation corresponding to each individual node (and each individual channel) as $\hat{\boldsymbol{\tau}}^n_{ij}$ goes to zero before local over-smoothing occurs in the $j$-th channel on a node $i$. 

\paragraph{Residual connections}
Motivated by the success of residual neural networks (ResNets) \citep{resnet} in conventional deep learning, there has been many suggestions of adding residual connections to deep GNNs. An early example includes \citet{resGCN}, where the authors equip a GNN with a residual connection \cite{resnet}, i.e.,
\begin{equation}
\label{eq:resGCN}
\bX^n = \bX^{n-1} + \bF_{\theta^n}(\bX^{n-1},\cG).
\end{equation}
By instantiating the GNN in \eqref{eq:resGCN} with a GCN, this leads to major improvements over competing methods. 
Another example is GCNII \citep{gcnii} where a scaled residual connection of the initial node features is added to every layer of a GCN,
\begin{equation}
\label{eq:gcnii}
\bX^n = \sigma\left[\left((1-\alpha_n)\hat{\bD}^{-\frac{1}{2}}\hat{\bA} \hat{\bD}^{-\frac{1}{2}}\bX^{n-1} + \alpha_n\bX^0\right) \left((1-\beta_n)\ident + \beta_n\bW^n \right) 
\right],
\end{equation}
where $\alpha_n,\beta_n \in [0,1]$ are fixed hyperparameters for all $n=1,\dots,N$. 
This allows for constructing very deep GCNs, outperforming competing methods on several benchmark tasks. Similar approaches aggregate not just the initial node features but all node features of every layer of a deep GNN at the final layer. Examples of such models include Jumping Knowledge Networks (JKNets) \citep{jknet} and Deep Adaptive Graph Neural Networks (DAGNNs) \citep{dagnn}

\subsection{Empirical evaluation}
\begin{figure}[ht!]
\begin{minipage}{\textwidth}
\includegraphics[width=.33\textwidth]{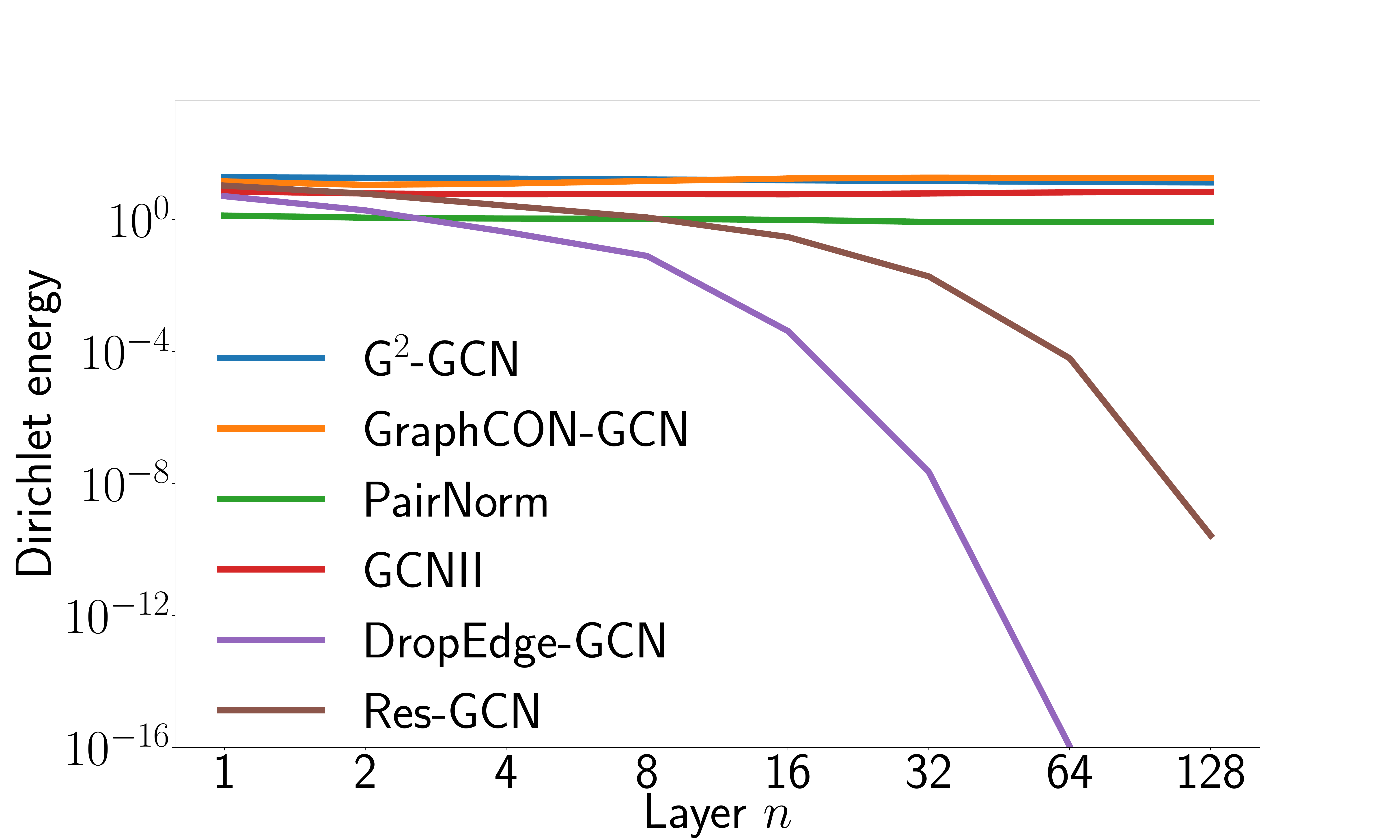}
\includegraphics[width=.33\textwidth]{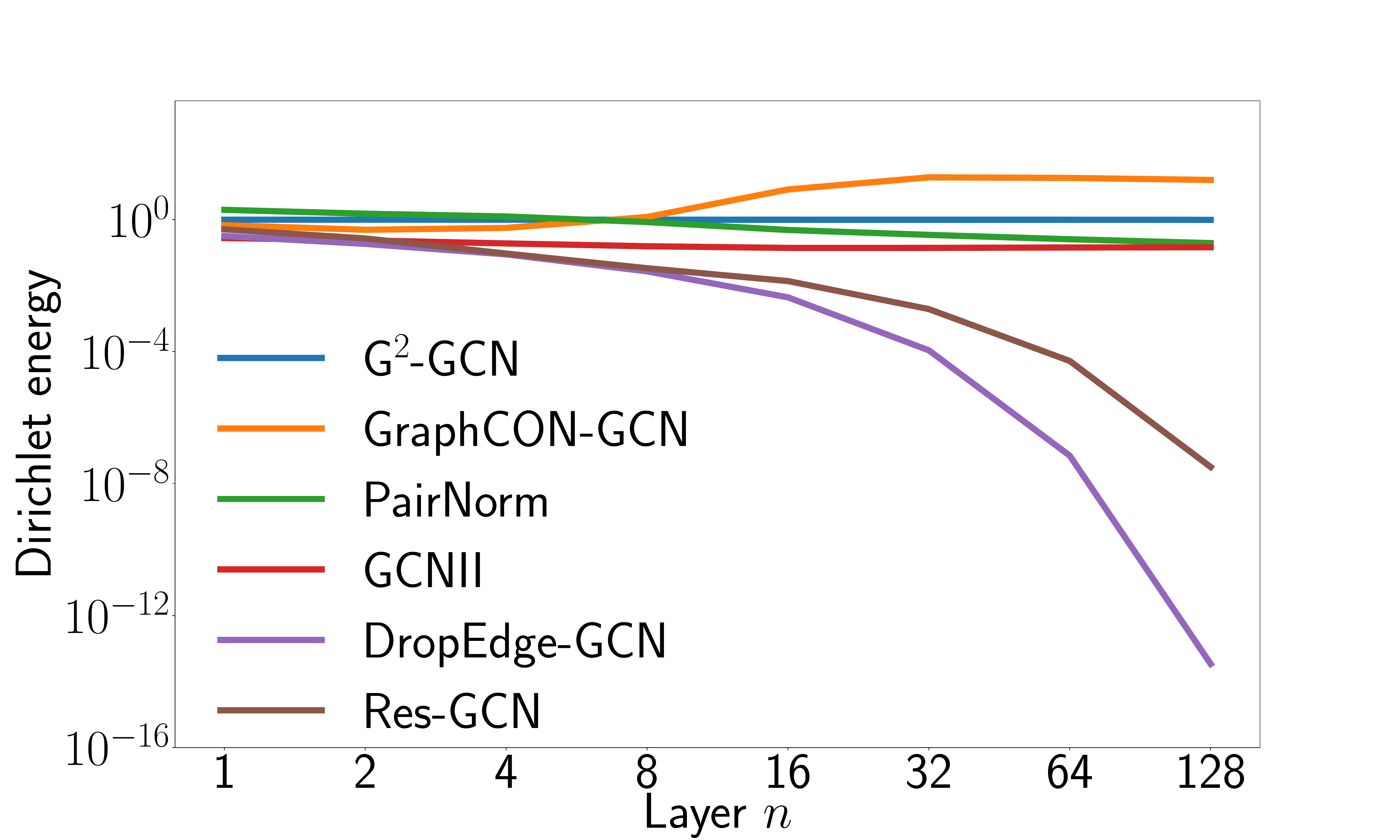}
\includegraphics[width=.33\textwidth]{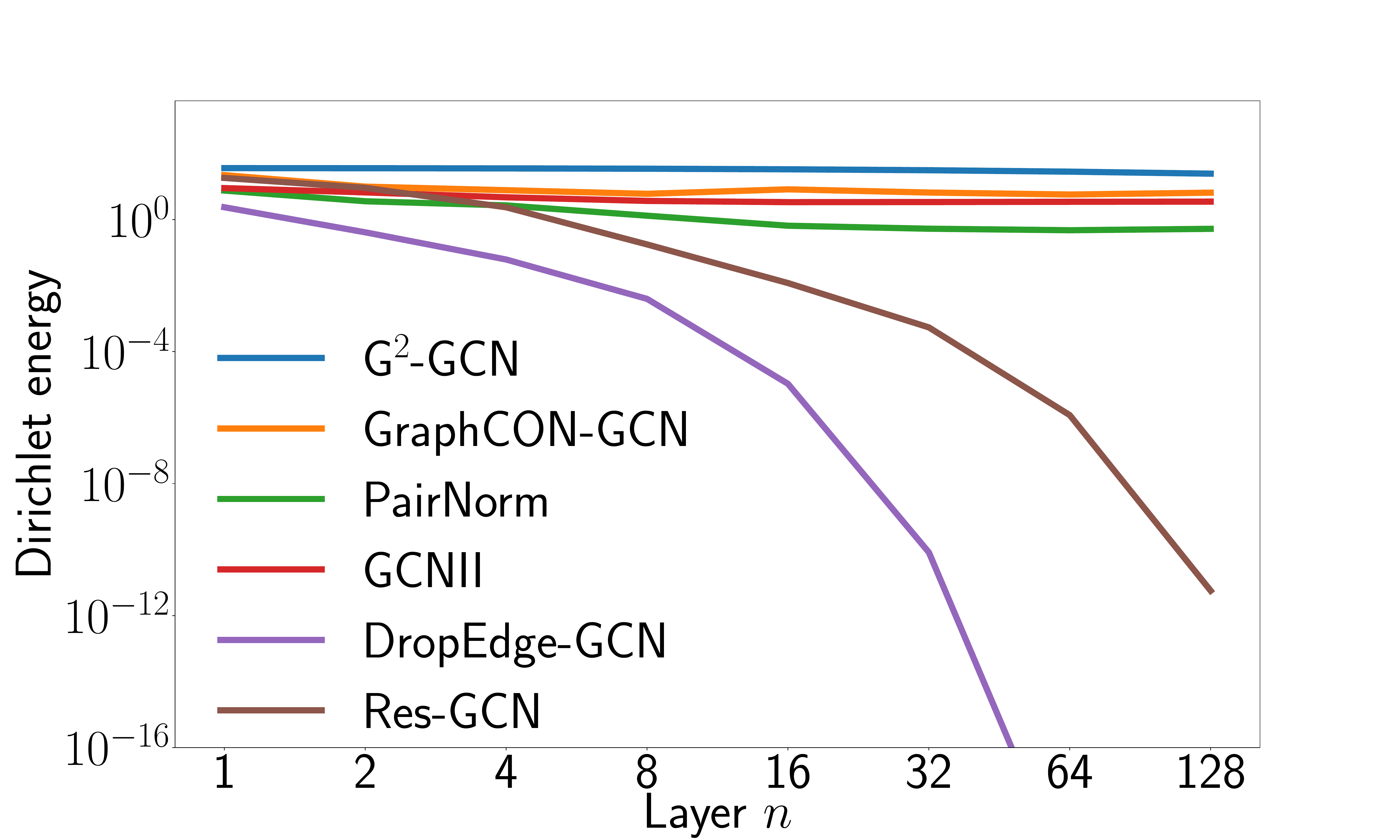}
\end{minipage}
\caption{Layer-wise Dirichlet energy of hidden node features propagated through \textbf{G}$^2$-GCN, GraphCON-GCN, PairNorm, GCNII, DropEdge-GCN and Res-GCN on three different graphs, i.e., \textbf{(left)} small-scale Texas graph, \textbf{(middle)} medium-scale Cora citation network, (\textbf{right}) large-scale Facebook (Cornell5) network.}
\label{fig:os_mitigation}
\end{figure}

In order to evaluate the effectiveness of the different methods that have been suggested to mitigate over-smoothing in deep GNNs, we follow the experimental set-up of section \ref{sec:measures}. To this end, we choose two representative methods of each of the different strategies to overcome over-smoothing, namely DropEdge and PairNorm as representatives from ``normalization and regularization'' strategies, GraphCON and \textbf{G}$^2$ from ``change of GNN dynamics'', and Residual GCN (Res-GCN) and GCNII from ``residual connections''. We consider the same three different graphs as in section \ref{sec:measures}, namely small-scale Texas, medium-scale Cora and larger-scale Cornell5 graph. Since we are only interested in the qualitative behavior of the different methods, we fix one node-similarity measure, namely the Dirichlet energy. Thereby, we can see in \fref{fig:os_mitigation} that DropEdge-GCN and Res-GCN suffer from an exponential convergence of the layer-wise Dirichlet energy to zero (and thus from over-smoothing) on all three graphs.
In contrast to that, all other methods we consider here mitigate over-smoothing by keeping the layer-wise Dirichlet energy approximately constant. 

\section{Risk of sacrificing expressivity to mitigate over-smoothing}
Since several of the previously suggested methods designed to mitigate over-smoothing successfully prevent the layer-wise Dirichlet energy (and other node-similarity measures) from converging exponentially fast to zero, it is natural to ask if this is already sufficient to construct (possibly very) deep GNNs which also efficiently solve the learning task at hand.
To answer this question, we start by constructing a deep GCN which keeps the Dirichlet energy constant while at the same time its performance on a learning task is as poor as a standard deep multi-layer GCN. 

\begin{figure}[ht!]
\includegraphics[width=.5\textwidth]{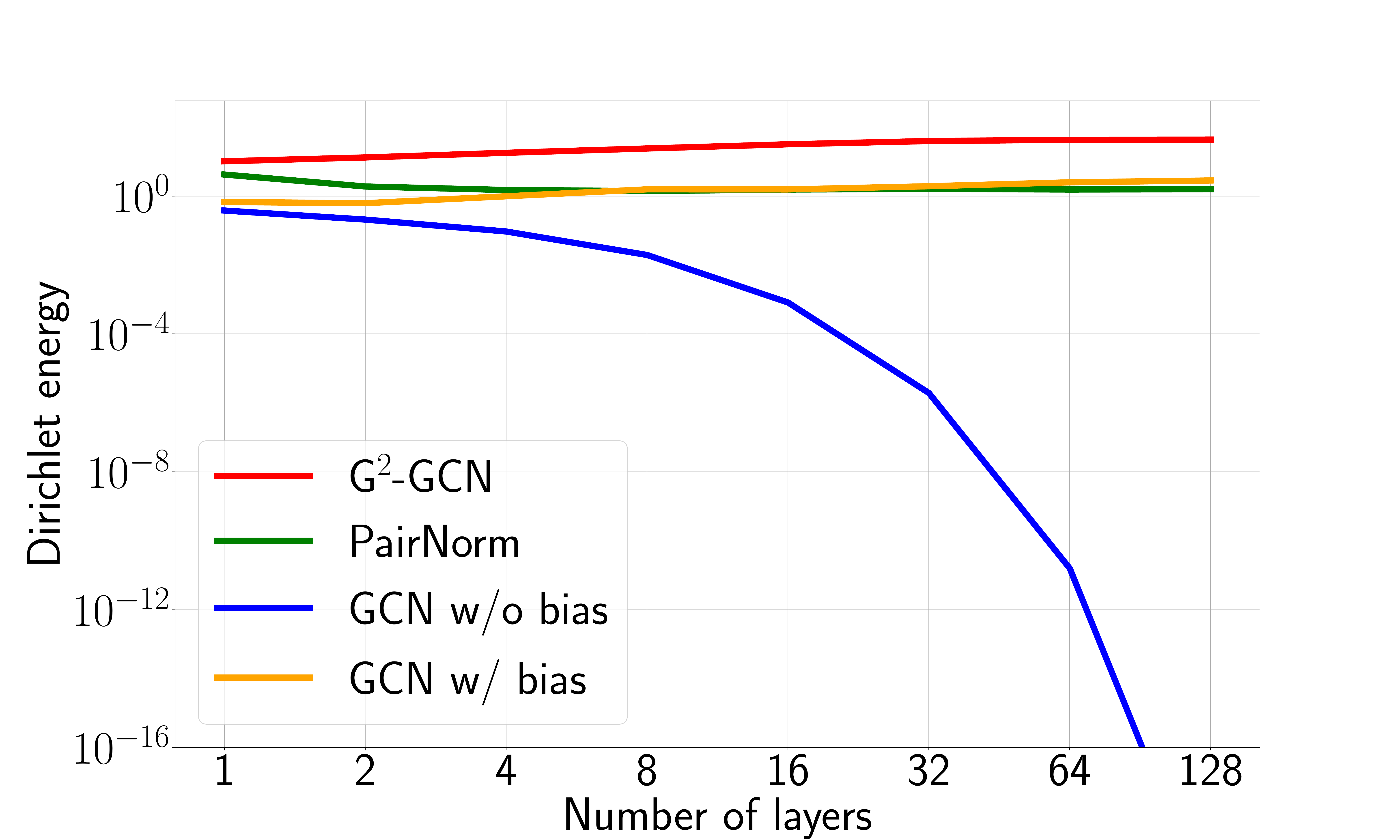}
\includegraphics[width=.5\textwidth]{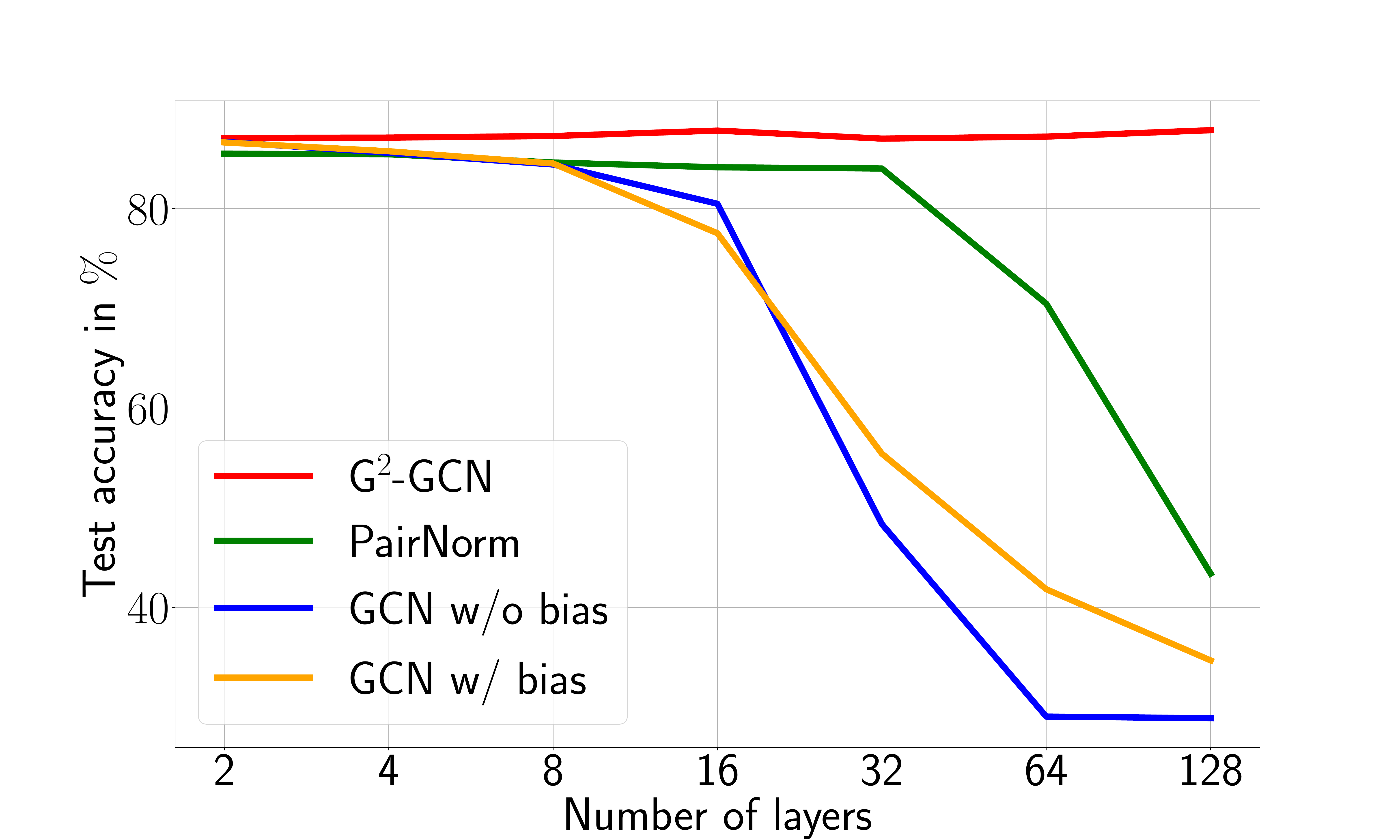}
\caption{Trained \textbf{G}$^2$-GCN, PairNorm, GCN with bias and GCN without bias on the fully-supervised Cora graph dataset using the pre-defined $10$ splits from \citet{geom_gcn}, showing two different measures for increasing number of layers ranging from $1$ to $128$:
(\textbf{left}) Dirichlet energy of the layer-wise node features, (\textbf{right}) test accuracies.}
\label{fig:os_pitfall_dirichlet}
\end{figure}

It turns out that simply adding a bias vector to a deep GCN with shared parameters among layers, i.e.,
\begin{equation*}
    \bX^n = \sigma(\hat{\bD}^{-\frac{1}{2}}\hat{\bA} \hat{\bD}^{-\frac{1}{2}}\bX^{n-1}\bW + \bb), \quad \forall n=1,\dots,N,
\end{equation*}
with weights $\bW \in \mathbb{R}^{m\times m}$ and bias $\bb \in \mathbb{R}^{m}$, is sufficient for the optimizer to keep the resulting layer-wise Dirichlet energy of the model approximately constant. This can be seen in \fref{fig:os_pitfall_dirichlet} where the layer-wise Dirichlet energy is shown (among others) for a standard GCN as well as a GCN with an additional bias term after training on the Cora graph dataset in the fully supervised setting. We observe that while the Dirichlet energy converges exponentially fast to zero for the standard GCN, simply adding a bias term results in an approximately constant layer-wise Dirichlet energy. Moreover, \fref{fig:os_pitfall_dirichlet} shows the test accuracy of the same models for different number of layers. We can see that both the standard GCN as well as the GCN with bias vector suffer from a significant decrease of performance for increasing number of layers. Interestingly, while GCN with bias keeps the Dirichlet energy perfectly constant and GCN without bias exhibits a Dirichlet energy converging exponentially fast to zero, both models suffer similarly from drastic impairment of performance (in terms of test accuracy) for increasing number of layers. We thus observe that simply constructing a deep GNN that keeps the node-similarity measure constant (around $1$) is not sufficient in order to successfully construct deep GNNs.

This observation is further supported in \fref{fig:os_pitfall_dirichlet} by looking at the Dirichlet energy of the PairNorm method which behaves similarly to the Dirichlet energy of GCN with bias, i.e., approximately constant around $1$. However, the performance in terms of test accuracy on the Cora graph dataset drops exponentially after using more than $32$ layers. Interestingly, \textbf{G}$^2$-GCN exhibits an approximately constant layer-wise Dirichlet energy and at the same time does not decrease its performance by increasing number of layers. In fact the performance of \textbf{G}$^2$-GCN increases slightly by increasing number of layers.

Therefore, we argue that solving the over-smoothing issue defined in Definition \ref{def:oversmoothing} is necessary in order to construct well performing deep GNNs. Otherwise the network is not able to learn any meaningful function defined on the graph. However, as can be seen from this experiment it is not sufficient.
Therefore, based on this experiment we conclude that a major pitfall in designing deep GNNs that mitigate over-smoothing is to sacrifice the expressive power of the GNN only to keep the node-similarity measure approximately constant. In fact, based on our experiments, only \textbf{G}$^2$ (among the considered models here) fully mitigates the over-smoothing issue by keeping the node-similarity measure approximately constant, while at the same time increasing its expressive power for increasing number of layers.

\section{Extension to continuous-time GNNs}
A rapidly growing sub-field of graph representation learning deals with GNNs that are continuous in depth. This is performed by formulating the message-passing propagation in terms of graph dynamical systems modelled by (neural \citep{node}) Ordinary Differential Equations (ODEs) or Partial Differential Equations (PDEs), i.e., message-passing framework \eqref{eq:mp}, where the forward propagation is modeled by a differential equation:
\begin{equation}
\label{eq:ct_mp}
    \bX^\prime(t) = \sigma(\bF_{\theta}(\bX(t),\cG)), 
\end{equation}
with $\bX(t)$ referring to the node features at time $t\geq0$. Different choices of the vector field (i.e., right-hand side of \eqref{eq:ct_mp}) yields different architectures. Moreover, we note that the right-hand side in \eqref{eq:ct_mp} can potentially arise from a discretization of a differential operator defined on a graph leading to a PDE-inspired architecture. 
We refer to this class of graph-learning models as \emph{continuous-time GNNs}.
Early examples of continuous-time GNNs include Graph Neural Ordinary Differential Equations (GDEs) \citep{gde} and Continuous Graph Neural Networks (CGNN) \citep{cgnn}. More recent examples include Graph-Coupled Oscillator Networks (GraphCON) \citep{graphcon}, Graph Neural Diffusion (GRAND) \citep{grand}, Beltrami Neural Diffusion (BLEND) \citep{blend}, Neural Sheaf Diffusion (NSD) \citep{sheaf}, and Gradient Glow Framework (GRAFF) \citep{grad_flow}.
Based on this framework, we can easily extend our definition of over-smoothing \ref{def:oversmoothing} to continuous-time GNNs, by defining over-smoothing as the exponential convergence in time of a node-similarity measure. More concretely, we define it as follows.
\vspace{1em}
\begin{defi}[Continuous-time over-smoothing]
\label{def:ct_oversmoothing}
Let $\cG$ be an undirected, connected graph and $\bX(t) \in \mathbb{R}^{v \times m}$ denote the hidden node features of a continuous-time GNN \eqref{eq:ct_mp} at time $t\geq0$ defined on $\cG$. Moreover, $\mu$ is a node-similarity measure as of Definition \ref{def:oversmoothing}.
We then define over-smoothing with respect to $\mu$ as the exponential convergence in time of the node-similarity measure $\mu$ to zero, i.e.,
\begin{enumerate}
    \item[] $\mu(\bX(t)) \leq C_1e^{-C_2 t}$, for $t\geq 0$ with some constants $C_1,C_2>0$.
\end{enumerate}
\end{defi}

\section{Conclusion}
Stacking multiple message-passing layers (i.e., a deep GNN) is necessary in order to effectively process information on relational data where the underlying computational graph exhibits (higher-order) long-range interactions. This is of particular importance for learning heterophilic graph data, where node labels may differ significantly from those of their neighbors.
Besides several other identified problems (e.g., over-squashing, exploding and vanishing gradients problem), the over-smoothing issue denotes a central challenge in constructing deep GNNs. 

Since previous work has measured over-smoothing in various ways, we unify those approaches by providing an axiomatic definition of over-smoothing through the layer-wise \emph{exponential convergence} of \emph{similarity measures} on the node features. Moreover, we review recent measures for over-smoothing and, based on our definition, rule out the commonly used MAD in the context of measuring over-smoothing. Additionally, we test the qualitative behavior of those measures on three different graph datasets, i.e., small-scale Texas graph, medium-scale Cora graph, and large-scale Cornell5 graph, and observe an exponential convergence to zero of all measures for standard GNN models (i.e., GCN, GAT, and GraphSAGE). 
We further review prominent approaches to mitigate over-smoothing and empirically test whether these methods are able to successfully overcome over-smoothing by plotting the layer-wise Dirichlet energy on different graph datasets.   

We conclude by highlighting the need for balancing the ability of models to mitigate over-smoothing, but without sacrificing the expressive power of the underlying deep GNN. This phenomenon was illustrated by the example of a simple deep GCN with shared parameters among all layers as well as a bias, where the optimizer rapidly finds a state of parameters during training that leads to a mitigation of over-smoothing (i.e., approximately constant Dirichlet energy). However, in terms of performance (or accuracy) on the Cora graph-learning task, this model fails to outperform its underlying baseline (i.e., same GCN model without a bias) which suffers from over-smoothing. This behavior is further observed in other methods that are particularly designed to mitigate over-smoothing, where the over-smoothing measure remains approximately constant but at the same time the accuracy of the model drops significantly for increasing number of layers. However, we also want to highlight that there exist methods that are able to mitigate over-smoothing while at the same time maintaining its expressive power on our task, i.e., \textbf{G}$^2$.
We thus conclude that \emph{mitigating over-smoothing is only a necessary condition}, among many others, for building deep GNNs, while a particular focus in designing methods in this context has to be on the maintenance or potential enhancement of the expressive power of the underlying model.

\section*{Acknowledgement}
The authors would like to thank Dr. Petar Veličković (DeepMind/University of Cambridge) for his insightful feedback and constructive suggestions.

\bibliography{refs}
\bibliographystyle{ref_style}

\end{document}